\documentclass[letterpaper, 10 pt, conference]{ieeeconf}  % Comment this line out if you need a4paper

\IEEEoverridecommandlockouts                              % This command is only needed if 
\usepackage{caption}
\usepackage{graphicx} % for \graphicspath
\usepackage{subcaption}
\usepackage{url}
\usepackage{xcolor}
\usepackage{algorithmic}
\usepackage[linesnumbered,ruled,vlined]{algorithm2e}
\usepackage{amssymb,amsmath,comment,cite}
\usepackage{amsfonts}
\usepackage{xspace}
\usepackage{booktabs}
\usepackage{multicol}
\usepackage{multirow}
\usepackage{cleveref}

\begin{document}
%%%%%%%%%%%%%%%%%%%%%%%%%%%%%%%%%   Title   %%%%%%%%%%%%%%%%%%%%%%%%%%%%%%%%%
\newtheorem{theorem}{Theorem}
\newtheorem{lemma}{Lemma}
\newtheorem{definition}{Definition}
\newcommand{\chutemapping}{task mapping\xspace}
\newcommand{\chutemappings}{task mappings\xspace}
\newcommand{\Chutemapping}{Task mapping\xspace}
\newcommand{\Chutemappings}{Task mappings\xspace}
\newcommand{\ChuteMapping}{Task Mapping\xspace}
\newcommand{\ChuteMappings}{Task Mappings\xspace}
\newcommand{\mysubsubsection}[1]{\noindent \textbf{#1}}

% For maps
\newcommand{\sortationSmallDense}{\textit{sortation-33-57-253}\xspace}
\newcommand{\sortationSmallSparse}{\textit{sortation-33-57-105}\xspace}
\newcommand{\sortationLargeDense}{\textit{sortation-50-86-703}\xspace}
\newcommand{\sortationLargeSparse}{\textit{sortation-50-86-325}\xspace}

%In case you encounter the following error:
%Error 1010 The PDF file may be corrupt (unable to open PDF file) OR
%Error 1000 An error occurred while parsing a contents stream. Unable to analyze the PDF file.
%This is a known problem with pdfLaTeX conversion filter. The file cannot be opened with acrobat reader
%Please use one of the alternatives below to circumvent this error by uncommenting one or the other
%\pdfobjcompresslevel=0
%\pdfminorversion=4

% See the \addtolength command later in the file to balance the column lengths
% on the last page of the document

% The following packages can be found on http:\\www.ctan.org
%\usepackage{graphics} % for pdf, bitmapped graphics files
%\usepackage{epsfig} % for postscript graphics files
%\usepackage{mathptmx} % assumes new font selection scheme installed
%\usepackage{times} % assumes new font selection scheme installed
%\usepackage{amsmath} % assumes amsmath package installed
%\usepackage{amssymb}  % assumes amsmath package installed

\title{\LARGE \bf
Destination-to-Chutes Task Mapping Optimization for\\Multi-Robot Coordination in Robotic Sorting Systems}

\author{
Yulun Zhang$^{1}$, Alexandre O. G. Barbosa$^{2}$, Federico Pecora$^{2}$, Jiaoyang Li$^{1}$% <-this % stops a space
% \thanks{*This work was not supported by any organization}% <-this % stops a space
\thanks{$^{1}$Yulun Zhang and Jiaoyang Li are with the Robotics Institute, Carnegie Mellon University. 
{\tt\small \{yulunzhang,jiaoyangli\}@cmu.edu}}%
\thanks{$^{2}$Alexandre O. G. Barbosa and Federico Pecora are with Amazon Robotics. 
{\tt\small \{aormiga,fpecora\}@amazon.com}}%
\thanks{This work is done during Yulun's internship at Amazon Robotics.}
}
% \author{Anonymous}
\maketitle
\thispagestyle{empty}
\pagestyle{empty}

%%%%%%%%%%%%%%%%%%%%%%%%%%%%%%%%%%%%%%%%%%%%%%%%%%%%%%%%%%%%%%%%%%%%%%%%%%%%%%%%
\begin{abstract}

We study optimizing a \emph{destination-to-chutes \chutemapping} to improve throughput in Robotic Sorting Systems (RSS), where a team of robots sort packages on a sortation floor by transporting them from induct workstations to eject chutes based on their shipping destinations (e.g. Los Angeles or Pittsburgh).
% To improve the throughput of RSS, prior works have studied many sub-problems, such as the coordination of hundreds of robots and the assignment of tasks to robots. However, robots' tasks in prior works are given directly as targets sampled from a known distribution.
% Prior works have studied the underlying Lifelong Multi-Agent Path Finding (MAPF) problem to improve the throughput of RSS. MAPF involves moving robots from starts to goals without collisions. Lifelong MAPF constantly assigns new goals to robots. 
% To determine which chutes to drop a package, we map the destination of the packages to chutes with a pre-defined \emph{destination-to-chutes} \chutemapping. The targets of the robots is traveling to one of the endpoints around the mapped chutes.
The destination-to-chutes \chutemapping is used to determine which chutes a robot can drop its package.
% While robots are transporting packages, they need to travel to a chute.
% In RSS, robots' tasks are given as packages. 
% To determine the target of a package, the system maps the destination of the package to a chute with a pre-defined \emph{destination-to-chutes} \chutemapping. 
Finding a high-quality \chutemapping is challenging because of the complexity of a real-world RSS. First, optimizing \chutemapping is interdependent with robot target assignment and path planning.
% Second, once enough packages are dropped into a chute, that chute shall be closed for the human workers on the lower floor to transport the packages for delivery.
Second, chutes will be \emph{CLOSED} for a period of time once they receive sufficient packages to allow for downstream processing.
% Third, the quality of task mapping is correlated with the efficiency of the human workers on the lower floor. If chutes of the same destinations are scattered, it is more time-consuming for human workers to handle the packages.
Third, \chutemapping quality directly impacts the downstream processing, as scattered chutes for the same destination increase package handling time. 
% These real-world factors make it difficult to define an accurate metric to evaluate the quality of a \chutemapping.
In this paper, we first formally define \chutemappings and the problem of \ChuteMapping Optimization (TMO).
% and show that the quality of a \chutemapping can significantly impact the traffic congestion of the robots.
We then present a simulator of RSS to evaluate \chutemappings. 
% We then discuss two heuristic functions that are relevant to high-quality \chutemappings and present two greedy algorithms to generate \chutemappings based on the heuristics.
We then present a simple TMO method based on the Evolutionary Algorithm and Mixed Integer Linear Programming, demonstrating the advantage of our optimized \chutemappings over the greedily generated ones in various RSS setups with different map sizes, numbers of chutes, and destinations. 
Finally, we use Quality Diversity algorithms to analyze the throughput of a diverse set of \chutemappings.
Our code is available online at \url{https://github.com/lunjohnzhang/tmo_public}.
% We additionally show the correlation between \chutemapping and other operational decisions in RSS, including path planning and target assignment. We demonstrate the advantage of our optimized mapping while being coupled with other operational decisions.
\end{abstract}

\section{Introduction}

We study \ChuteMapping Optimization (TMO), the problem of optimizing the \chutemapping in Robotic Sorting Systems (RSS). With the flourishing of e-commerce and online shopping, the demand for more efficient supply chains for delivering packages as quickly as possible is growing rapidly. While preparing the packages for shipping, e-commerce stakeholders need to sort packages based on their shipping destinations.
To fulfill the growing demand, major stakeholders such as Amazon~\cite{shen_multi-agent_2023,AR_RSS_FOX2019}, JD.com~\cite{JD2023Sorting}, Deppon Express~\cite{zou_robotic_2021}, and Shentong Express~\cite{Zheng2017Shengtong} use RSS with hundreds and thousands of robots to sort packages. Compared to traditional sorting systems based on conveyor belts~\cite{rohrer_simulation_1995,mcwilliams_parcel_2005} and Robotic Mobile Fulfillment Systems (RMFS)~\cite{Li2020LifelongMP,liItemStorageAssignment2024,barnhartRoboticWarehousingOperations2024,caiCollaborativeOptimizationStorage2021,kimItemAssignmentProblem2020,azadehRobotizedAutomatedWarehouse2019} used for storing goods, RSS is relatively new with many unaddressed research questions.

\begin{figure}[!t]
    \centering
    \includegraphics[width=1\linewidth]{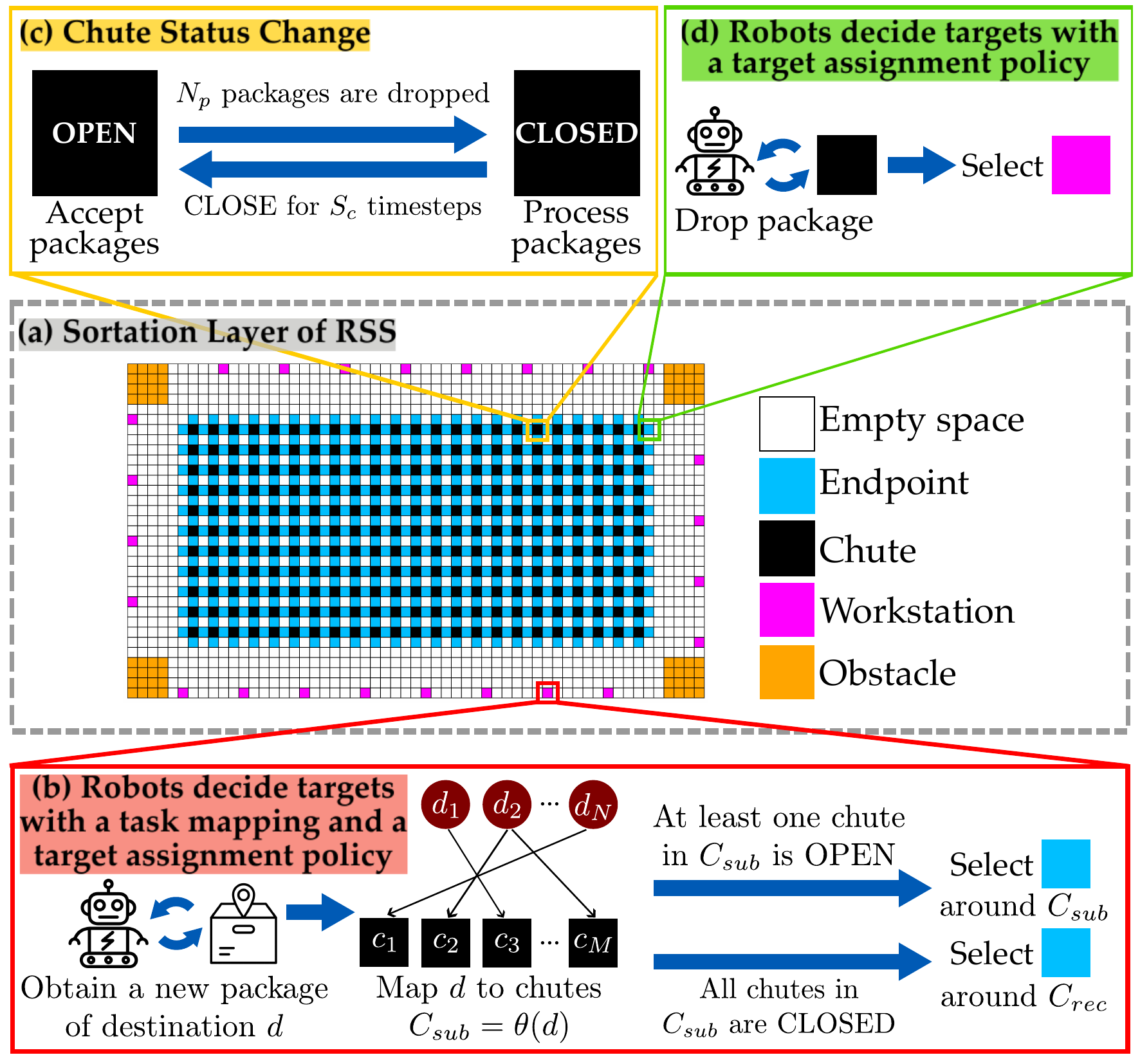}
    \caption{Overview of our RSS simulator. (a) Map of the sortation floor of the RSS. 
    % Robots repeatedly pick up packages at the workstations (pink), and move to the endpoints (blue) in the middle to drop the packages into the chutes (black). All spaces except for the chutes and obstacles are traversable. 
    (b) At the workstations, each robot picks up a package and decides the target endpoint with a task mapping and a TA policy. (c) Status change of the chutes. Packages can be dropped into \emph{OPEN} chutes. (d) At the endpoints, each robot drops off the package and decides the next target workstation with a TA policy.}
    % \Description[TODO]{todo}
    \label{fig:RSS-sim}
\end{figure}

% Highlight in the intro that we are talking about a real-world RSS, instead of a simulator.
% \Cref{fig:AR-sortation} shows robots transporting packages in a real-world RSS. 
% \Cref{fig:sortation-33-57} 
\Cref{fig:RSS-sim}(a) shows an example RSS map, where robots constantly move between workstations (pink) and endpoints (blue) to transport packages from workstations to chutes (black). When a robot picks up a package at a workstation, the RSS leverages a pre-defined \textit{destination-to-chutes} \chutemapping to map the destination of the package to a set of chutes. The robot selects a chute and an endpoint around it, 
moves to the endpoint, and drops the package into the chute. Finally, the robot moves to a workstation to pick up the next package. 
More details are explained in \Cref{sec:RSS}.

% \begin{figure}
%     \centering
%     \begin{subfigure}[b]{0.245\textwidth}
%       \centering
%       \includegraphics[width=1\textwidth]{figs/maps/AR_sortation_pic.png}
%       \caption{}
%       \label{fig:AR-sortation}
%     \end{subfigure}
%     \hfill
%     \begin{subfigure}[b]{0.23\textwidth}
%       \centering
%       \includegraphics[width=1\textwidth]{figs/maps/sortation_33_57.png}
%       \caption{}
%       \label{fig:sortation-33-57}
%     \end{subfigure}
%     \label{fig:front-fig}
%     \caption{(a) A real-world RSS of Amazon Robotics\cite{AR_RSS_FOX2019}. Robots are used to transport packages from workstations to chutes. (b) A 2D grid map of an RSS with workstations (pink), chutes (black), endpoints (blue), empty spaces (white), and obstacles (orange). Robots can traverse on spaces other than chutes and obstacles.}
% \end{figure}

% RSS leverages a pre-defined \textit{destination-to-chutes} \chutemapping to decide which chutes the packages should be dropped into. 
The quality of the \chutemapping significantly affects the movement patterns of the robots.
Since the volume (i.e., number of packages) of different destinations varies, the chutes that are mapped to high-volume destinations can be visited more frequently.
If chutes mapped to high-volume destinations are clustered together, robots would travel to endpoints around those chutes more frequently, potentially accumulating congestion and lowering throughput. Meanwhile, if high-volume destinations map to chutes that are close to the workstations, the travel distance of the robots can be reduced and thus improve throughput. 
% Therefore, it is important to optimize a high-quality \chutemapping, inducing balanced traffic and improving throughput.

% Therefore, optimizing a high-quality \chutemapping is essential for a high-throughput RSS.

However, TMO is a challenging problem because of the complex nature of an RSS. First, TMO is strongly correlated with other sub-problems in RSS, including (1) robot coordination, the problem of planning collision-free paths for the robots, and (2) target assignment (TA), the problem of selecting appropriate endpoints or workstations for robots to go to. 
TMO is correlated with these two problems because the movement of the robots is collectively determined by a \chutemapping, a robot coordination algorithm, and a TA policy.
Second, the capacity of the chutes is limited. Once a sufficient number of packages are dropped into a chute, it undergoes a \emph{status change}, transitioning status to be \emph{CLOSED} for a certain amount of time, during which no packages can be dropped. 
% Therefore, a high-quality \chutemapping needs to assign enough chutes to high-volume destinations. 
Third, the operation of RSS depends not only on the \emph{sortation floor} on which robots move, but also on the downstream process on a lower \emph{shipping floor} where the dropped packages are collected for delivery. If chutes of the same destination are clustered together, it would alleviate the workload on the shipping floor, which decreases the amount of time chutes are \emph{CLOSED} and improves throughput on the sortation floor. This trades off with the fact that clustered chutes of the same destination can impede the movement of the robots. These real-world factors make it challenging to not only solve TMO, but also to develop a simulator that can effectively evaluate a given \chutemapping.

% Talk about why doing these things are important
In this paper, we first formally define \ChuteMapping Optimization (TMO). To the best of our knowledge, very few prior works have studied TMO in the context of RSS.
% Our key insight is that \chutemappings determines the traffic pattern of the robots and thus the throughput of RSS. 
We then present a simulator that considers the real-world factors of an RSS to evaluate \chutemappings.
% We then discuss two heuristic functions that are relevant to the quality of a \chutemapping and propose two greedy algorithms to generate \chutemappings based on the two functions, respectively. 
We then present a simple optimization algorithm based on Evolutionary Algorithm (EA) and Mixed Integer Linear Programming (MILP) to solve TMO, showing that the optimized \chutemappings outperform greedily generated ones. Finally, we use Quality Diversity (QD) algorithms~\cite{justin_qd_2016} to conduct empirical analysis on a diverse set of \chutemappings.

% Based on this observation, we propose to optimize the \textit{capacity} of each chute, defined as the expected volume that the chute can handle without causing significant traffic congestion. We first develop a Mixed Integer Linear Programming (MILP) solver to generate the \chutemapping given the capacities of each chute and the expected volume of the packages for each destination. We then use CMA-ES~\cite{hansen2016cmaes}, a state-of-the-art single-objective derivative-free optimization algorithm, to search for the optimal chute capacities with the objective of maximizing the throughput. We compute throughput by leveraging a lifelong MAPF simulator. We will show in our experiment that our proposed approach outperforms the baseline analytical methods.

\section{Existing Research in RSS}

% Given the complex nature of the RSS, a number of sub-problems arise in such systems, including 
Prior works in RSS mainly focus on (1) robot coordination~\cite{Li2020LifelongMP}, the problem of planning collision-free paths for multiple robots, (2) target assignment~\cite{KouAAAI20}, the problem of selecting a target for each robot from a set of candidate targets, (3) package assignment to workstations~\cite{huang_flow-based_2023}, the problem of pre-sorting packages to different workstations based on their shipping destinations, 
% (4) package assignment to robots~\cite{huang_flow-based_2023}, the problem of deciding which robot transports which package, 
(4) layout optimization~\cite{zhao_robotic_2024}, the problem of optimizing the layout of the sortation floor of RSS to improve throughput, 
(5) RSS simulator development,
and (6) \ChuteMapping Optimization (TMO). In this section, we provide literature reviews of problems (1), (2), and (6), as they are explicitly considered in our RSS simulator.
We also discuss (5) because it is a challenging research problem orthogonal to the other sub-problems. Developing a simulator is necessary to either solve these problems or evaluate their solutions.

% Make this paragraph an overview, summarize the decisions we need to make in RSS, and then go to each decision
% Given the complex nature of the RSS, prior works have studied many sub-problems other than MAPF and TA.

% Another recent work has considered the problem of optimizing layouts of the sortation floor for RSS~\cite{zhao_robotic_2024}.

\mysubsubsection{Robot Coordination.}
% one paragraph on mapf
% many work focus on fulfillment center map
Many prior works have studied the coordination of robots in automated warehouses, especially in the underlying problem of lifelong Multi-Agent Path Finding (MAPF)~\cite{SternSoCS19,Li2020LifelongMP}. Lifelong MAPF aims at finding collision-free paths for a group of robots from their corresponding start to target locations. New targets are constantly assigned to robots. Previous studies in lifelong MAPF have improved throughput significantly by developing better lifelong MAPF algorithms~\cite{WangICAPS08,MaAAMAS17,okumura2019priority,okumura2023lacam,LiAAMAS20a,KouAAAI20,DamaniRAL21,Jiang2024Competition}, optimizing the physical layouts~\cite{zhangLayout23,ZhangNCA2023}, or designing virtual guidance for robots~\cite{Yu2023,ChenAAAI24,zhang2024ggo,ZangAAAI25}. However, most prior works in MAPF assume that the targets of the robots are either directly given~\cite{Jiang2024Competition} or sampled from a known distribution of targets~\cite{Li2020LifelongMP,zhangLayout23}, and most of these works focus on RMFS instead of RSS.
% In RSS, however, the targets of the robots are mapped from packages via a \chutemapping. In addition, most of these works focus on RMFS instead of RSS.
The operation research community has also attempted to tackle the robot coordination problem in RSS. One work~\cite{huang_flow-based_2023} applies multi-commodity network flow to determine the expected number of robots that can travel across the sortation floor.
% simultaneously solve three problems in RSS given a fixed \chutemapping: (1)  robots coordination, (2) the assignment of packages to workstations, and (3) the assignment of robots to workstations. The solution flow determines both the assignments as well as the expected number of robots that can travel across the sortation floor. However, 
% since it does not consider collisions between robots, it is non-trivial to use state-of-the-art lifelong MAPF algorithms to move the robots following the solution flow. In addition, 
% However, it only applies to small sorting systems with a map size of 19 $\times$ 20, 2 workstations, and 35 robots. 
Another work~\cite{zhao_robotic_2024} applies Rhythmic Control (RC), an autonomous vehicle management scheme, to coordinate robots in RSS. 
% However, they only compare with Cooperative A$^*$~\cite{SilverAIIDE05}, an incomplete and suboptimal MAPF algorithm based on single-agent A$^*$, instead of state-of-the-art planners.

\mysubsubsection{Target Assignment.}
% Look into OR research
The target assignment (TA) problem aims at assigning each robot to a target among the available options. For example, while sorting a package, a robot can go to any endpoints adjacent to chutes corresponding to the destination of the package. 
% Similarly, while picking up a new package, a robot can go to any workstation. 
An appropriately selected endpoint can potentially minimize traffic congestion or travel distance of the robot, improving throughput. Early works on classical assignment problems~\cite{Kuhn1955Hungarian,Gross1959THEBA} can be used to solve TA. 
In the MAPF community, one work~\cite{Li2020LifelongMP} uses a simple greedy TA policy that selects a target based on distances to the available targets and numbers of robots traveling to the available targets. 
Another work~\cite{KouAAAI20} proposes a min-cost max-flow framework to solve TA. 
% They also studied the TA and Path Finding (TAPF) problem, which aims at solving TA and MAPF at the same time, under the context of RSS. 
% However, instead of optimizing throughput, they optimize the idle time of the workstations, defined as the amount of time when no robots visit the workstations. While they claim idle time to be correlated with throughput, no justification is provided. 
% However, the improvement in idle time appears to be minimal compared with a variant of the classical assignment algorithm~\cite{Kuhn1955Hungarian}. 
Another branch of works~\cite{zou_robotic_2021,xu_assignment_2022} in the operation research community compares different handcrafted TA policies by using Queueing Networks to model RSS and estimate the throughput. 
% However, no automatic optimization techniques are developed, and \chutemappings are not considered. 

\mysubsubsection{Sorting Systems Simulation.}
% Prior works have attempted to build a simulator to model the behavior of different sorting systems. 
Some early works~\cite{rohrer_simulation_1995,mcwilliams_parcel_2005} have built simulators for traditional conveyor-based sorting systems. For RSS, recent works have modeled it using a queuing network~\cite{zou_robotic_2021,xu_assignment_2022}, a traffic flow network~\cite{huang_flow-based_2023}, or an estimation formula~\cite{zhao_robotic_2024}.
However, they do not explicitly model the movement and collision of the robots or the status change of the chutes. 
% One work proposes an estimation formula to approximate the throughput of RSS given the number of workstations and robots. However, it does not consider \chutemappings or the status change of the chutes.
One work~\cite{shen_multi-agent_2023} considers \chutemapping by modeling the maximum number of packages the sortation floor can handle, but it does not consider the movement and collision of the robots.
The MAPF community has simulated the movement of the robots in RSS using state-of-the-art lifelong MAPF algorithms~\cite{KouAAAI20,Li2020LifelongMP,Jiang2024Competition}, but they do not consider the status change of the chutes or movement of the human workers on the shipping floor. 
% In addition, all the above works do not consider \chutemappings.

\mysubsubsection{\ChuteMapping Optimization.}
% task assignment,etc
% highlight: we use a more accurate simulator, instead of using a simplified flow model, etc
% Another work~\cite{boysen_robotized_2023} considers a different type of sorting system used for order consolidation.
% develops MILP to model the sorting system and optimize for a \chutemapping. However, similar to Queueing Network approaches, they do not model the robots' movement and collisions. In addition, instead of optimizing for the throughput directly, they optimize approximated measures such as the point in time that the last package is transported to the goal. They also focus on small sorting systems with only 1 workstation, 20 chutes, and as many as 40 robots, while other works~\cite{Yu2023,Li2020LifelongMP} have used sortation maps of size 179 $\times$ 69 and as many as 1000 robots.
To the best of the authors' knowledge, very few works have attempted to develop automatic optimization methods to optimize \chutemappings.
One related work~\cite{shen_multi-agent_2023} optimizes a \emph{dynamic} \chutemapping by training a Multi-Agent Reinforcement Learning (MARL) policy to dynamically determine the number of chutes that should be mapped to each destination.
% based on the volume of the packages and movements of the robots on the sortation floor. 
They then use a handcrafted rule to generate \chutemappings based on the determined numbers. The trained MARL policy dynamically adapts the number of chutes every hour during the execution of RSS, making the optimized \chutemapping dynamic. Our work is different in several ways. 
First, we optimize the \chutemappings directly, instead of indirectly by optimizing the numbers of chutes for each destination. 
Second, we consider all chutes while optimizing \chutemappings, while they keep the mapping of part of the chutes fixed.
Third, we focus on optimizing \emph{static}, instead of dynamic, \chutemappings. In real-world RSS, it is challenging to change the \chutemapping frequently because of the downstream dependencies on the shipping floor.
Fourth, their simulator does not model the movement and collisions of the robots. In addition, we cannot compare with their approach because the handcrafted rule used to generate a \chutemapping based on the number of chutes assigned to each destination is not published. 

% The state-of-the-art analytical method for \chutemapping is based on network flow~\cite{}. By representing the robotic sorting system as a multi-commodity network flow problem, they attempt to simultaneously solve three problems: (1) the MAPF problem, (2) the assignment of packages to chutes (\chutemapping), and (3) the assignment of packages to robots. The solution flow determines both the assignments as well as the expected number of robots that can travel on each edge. During execution, they move the robots based on the corresponding flow and use pre-defined rules to resolve robot collisions. 

% However, similar to the other two categories of methods, the network-flow-based approach does not explicitly model collisions between robots. 

\section{Problem Definition}

We formally define \chutemappings 
% valid \chutemappings
% , the optimal \chutemapping, 
and the problem of \ChuteMapping Optimization (TMO).

% \begin{definition}[\ChuteMapping]
% \label{def:chute-map}
%     Given $M$ chutes $C = \{c_1, ..., c_M\}$ and $N$ destinations $D = \{d_1, ..., d_N\}$, a \chutemapping is a function $m: C \rightarrow D$, while ensuring that all destinations are covered: $\bigcup_{c \in C} \theta(c) = D$.
% \end{definition}

\begin{definition}[\ChuteMapping and TMO]
    Given $M$ chutes $C = \{c_1, ..., c_M\}$ and $N$ destinations $D = \{d_1, ..., d_N\}$ ($M \geq N$), a \chutemapping is a function $
    \theta: D \rightarrow \mathcal{P}(C)$, where 
    % $C$ and $D$ are the sets of all chutes and destinations with $|C| = M$ and $|D| = N$, respectively, and
    $\mathcal{P}$ is the power set of $C$. A \chutemapping is \emph{valid} if and only if (1) each destination is mapped to \emph{at least} 1 chute, and (2) each chute is mapped to \emph{exactly} 1 destination. A valid \chutemapping is \emph{optimal} if it maximizes the \emph{throughput} of the RSS, defined as the number of tasks finished by all robots per timestep. The problem of \ChuteMapping Optimization (TMO) attempts to search for the optimal \chutemapping.

\end{definition}

% We need to ensure that the RSS is capable of sorting packages of all destinations and all chutes are appropriately leveraged. In addition, we assume that each chute can accept packages of exactly 1 destination. Therefore, we further define the \emph{valid} \chutemappings.

% \begin{definition}[Valid \ChuteMapping]
%     A \chutemapping is valid if and only if (1) each destination is mapped to \emph{at least} 1 chute, and (2) each chute is mapped to \emph{exactly} 1 destination.
% \end{definition}

% We then define the problem of \ChuteMapping Optimization (TMO).

% \begin{definition}[\ChuteMapping Optimization (TMO)]
%     Given $M$ chutes $C = \{c_1, ..., c_M\}$ and $N$ destinations $D = \{d_1, ..., d_N\}$ with volumes $V = \{v_1, ...,v_N\}$, where $M \geq N$, as well as an objective function $f: \Theta \rightarrow \mathbb{R}$, where $\Theta$ is the space of all possible \chutemappings, the TMO problem searches for the best valid \chutemapping that maximizes the objective function.
% \end{definition}

We require $M \geq N$ because a valid \chutemapping requires each chute to be mapped to exactly 1 destination. This constraint cannot be satisfied if there are fewer chutes than destinations. In fact, in real-world RSS, it is common to have more chutes than destinations, because one destination in RSS may map to a range of shipping locations.
% We evaluate the throughput of a \chutemapping by using a RSS simulator, discussed in \Cref{sec:RSS}.

\section{Robotic Sorting System} \label{sec:RSS}

% In this section, we first present our simulator used to evaluate the throughput of a given \chutemapping. We then demonstrate the effect of \chutemapping on traffic congestion. We then discuss two heuristic functions that are relevant to high-quality \chutemappings.

% \subsection{RSS Simulator} \label{sec:RSS:sim}

We present our simulator used to evaluate the throughput of a given \chutemapping. \Cref{fig:RSS-sim} shows the overview.

\mysubsubsection{Robot Coordination.}
Our robot model follows the standard MAPF model~\cite{SternSoCS19}. Robots move on a 4-connected grid graph $G(V, E)$ exemplified in \Cref{fig:RSS-sim}(a), where each cell represents a vertex. We discretize time into timesteps, and each robot can either move to an adjacent vertex or wait at the current vertex at each timestep. Two robots collide if they move to the same vertex or swap vertices at the same timestep. We use PIBT~\cite{okumura2019priority} as the robot coordination algorithm in our RSS simulator because it can coordinate thousands of robots within seconds~\cite{Jiang2024Competition}.
% in maps of size as large as 500 $\times$ 140.

\mysubsubsection{Target Assignment.}
We follow the previous work~\cite{Li2020LifelongMP} to use a simple greedy TA policy.
% that selects a target from available targets based on (1) the distances to the available targets, and (2) the numbers of robots traveling to the available targets. 
Specifically, given a set of available targets $T \subseteq V$, a robot selects the target $g \in T$ that minimizes $L(v, g) + \alpha N_r(g)$, where 
$v \in V$ is the current vertex of the robot,
$L: V \times V \rightarrow \mathbb{R}$ computes the shortest path length between two vertices in $G$,
$N_r: T \rightarrow \mathbb{N}$ returns the number of robots traveling to each target in $T$,  and $\alpha$ is a hyperparameter. We consider the workstations and endpoints around the \emph{OPEN} chutes as the targets. The function $N_r$ is an estimate of the expected traffic congestion around each target. Intuitively, the greedy TA policy selects the target by balancing the travel distances of the robots and the traffic congestion around the available targets. In all our simulator, we follow the previous work~\cite{Li2020LifelongMP} to use $\alpha = 8$.

\mysubsubsection{Picking and Dropping Packages.}
As shown in \Cref{fig:RSS-sim}(b), each robot picks up a package at a workstation and decides on an endpoint to go to.
% shows the process of picking up a package and determining the target endpoint. 
Given a \chutemapping $\theta$, each robot picks up a package of destination $d$ and maps $d$ to a set of chutes $C_{sub} = \theta(d)$. If at least one chute in $C_{sub}$ is \emph{OPEN}, the robot selects one of the endpoints around chutes in $C_{sub}$ based on the TA policy. If all chutes in $C_{sub}$ are \emph{CLOSED}, robots must travel to a set of \emph{recirculation chutes} $C_{rec}$. Packages that go to $C_{rec}$ will be redirected to the workstations to be sorted again. A package is \emph{sorted} if it is correctly dropped into a non-recirculation chute. Otherwise, it is \emph{recirculated}. 
While computing throughput, we ignore the recirculated packages.
% We evaluate the performance of RSS by \emph{throughput}, defined as the number of packages sorted per timestep.
After moving to the selected endpoint and dropping the package, the robot uses the TA policy to select the next target workstation, as shown in \Cref{fig:RSS-sim}(d). We optimize which chutes are assigned to $C_{rec}$ by considering ``recirculation'' as an additional destination.

% \color{red}{Mention how to change the definition of chute mapping to incorporate recirculation chute.}
\mysubsubsection{Chute Status Change.}
\Cref{fig:RSS-sim}(c) shows the status change of the chutes. Initially, all chutes are \emph{OPEN}. After $N_p$ packages are dropped into a chute $c$, it is \emph{CLOSED} as workers on the lower shipping floor need to process the packages. $c$ must remain \emph{CLOSED} for $S_c$ timesteps.
Since modeling the movement of human workers explicitly on the shipping floor overkills the simulator, we greedily estimate $S_c$ using a handcrafted equation.
We compute $S_c$ based on how clustered chutes of the same destination are on the sortation floor. Specifically, suppose $d_c$ is the destination of $c$, we first obtain $C_{sub\_c} = \theta(d_c)$, the set of chutes mapped to $d_c$. To quantify how clustered chutes in $C_{sub\_c}$ are, we compute the \emph{centroid} of $C_{sub\_c}$ by averaging their coordinates. We then compute the \emph{centroid distance} $x_c$ as the average Euclidean distance between every chute in $C_{sub\_c}$ and the centroid. 
To compute $S_c$, we first use a monotonically increasing function $s: \mathbb{R} \rightarrow \mathbb{R}$ to compute the \emph{minimal} time the chute must be \emph{CLOSED}.
The more scattered the chutes in $C_{sub\_c}$ are, the more time workers on the shipping floor need to process the packages, the longer the chutes must remain \emph{CLOSED}. Furthermore, the process time of the packages incurs stochasticity, so we sample from an exponential distribution $\epsilon \sim Exp(\beta)$ for an \emph{additional} amount of package processing time, resulting in a total \emph{CLOSED} time of $S_c = \lfloor s(x_c) + \epsilon \rfloor$. We use $N_p = 50$, $s = 2x_c^2 + 50$, and $\beta = 100$, modeling a minimal package processing time of 50 timesteps, and an average additional time of 100 timesteps.
% \footnote{We have no access to the real-world data of package processing time,}.

\mysubsubsection{Distribution of Destinations.}
The distribution of the destination of the packages significantly affects the resulting \chutemappings. For example, high-volume destinations (e.g., Los Angeles) should be mapped to more chutes than low-volume ones (e.g., Pittsburgh). Usually, packages of a small number of high-volume destinations consist of a large portion of all packages.
Therefore, we adapt the 7:2:1 distribution used in RMFS from the previous works~\cite{zhang2016_721_dist,Tang2024CachingAugmentedLM} to RSS, where 70\%, 20\%, and 10\% of the destinations are sampled 10\%, 20\%, and 70\% of the times, respectively.

% The impact of \chutemapping on robots' movement

% Centroid distance and min dist to workstations

\section{Task Mapping Optimization} \label{sec:tmo}

% \textcolor{red}{Change section A in a way that the greedy optimization methods are used for initialization of EA.}

Given that the RSS simulator is a non-differentiable black-box function, 
we present an automatic \chutemapping optimization method based on the $(1+\lambda)$ Evolutionary Algorithm (EA)~\cite{thomas1997EA}.
% , a population-based single-objective optimizer capable of optimizing non-differentiable black-box functions. 
After initializing a population of $\lambda$ \chutemappings (\Cref{subsec:ea-init}), EA maintains the population by iteratively undergoing selection, mutation, and evaluation to search for high-throughput \chutemappings. In each iteration, the one \chutemapping with the highest throughput is selected for mutation (\Cref{subsec:ea-mutate}). We evaluate a given \chutemapping by running $N_e$ simulation in our RSS simulator for $N_T$ timesteps and compute the average throughput as the objective.
% to ensure that the evaluated \chutemappings are valid, 
We use a MILP solver to enforce validity constraints on the mutated \chutemappings (\Cref{subsec:ea-milp}). We stop EA when we have evaluated $N_{eval}$ \chutemappings.

\subsection{Initialization} \label{subsec:ea-init}

Having a good initial set of solutions is critical for EA to search for high-throughput \chutemappings. 
% However, due to the limited computational budget, it is impractical to use the RSS simulator as the objective while generating the initial \chutemappings. 
Our EA utilizes three initialization strategies based on different approximate objectives that are correlated to throughput.
% Therefore, we present three greedy optimizers, each optimizing an objective that approximates throughput in a RSS, to generate the initial solutions. 

\mysubsubsection{Sampling from the Distribution of Destinations.} Intuitively, the more packages a destination is expected to have, the more chutes are required to absorb those packages. Therefore, a basic initialization technique is sampling a destination from the distribution of destinations for each chute. This method ensures the number of chutes assigned to each destination is proportional to the volume of the destination, but it does not explicitly optimize the mapping between chutes and destinations.

% Heuristics for a High-quality \ChuteMapping
\mysubsubsection{Min-dist Greedy Initialization.} 
This technique generates a \chutemapping 
% The first greedy optimizer optimizes the movement of the robots on the sortation floor 
by greedily assigning high-volume destinations  to chutes closer to workstations. Chutes with high-volume destinations are more frequently visited by the robots. Therefore, placing such chutes closer to the workstations allows more robots to travel shorter distances, thereby potentially improving throughput. This optimizer focuses on optimizing the movement of the robots on the sortation floor.

% We propose this optimizer based on several key observations. First, high-volume chutes are visited more frequently by the robots. Therefore, by minimizing their path lengths to the workstations, we reduce the travel distance of the robots, potentially allowing them to sort more packages within the same timesteps. Second, our target assignment policy prefers robots to go to closer workstations upon picking up a new package at the workstations. Therefore, for each chute, we minimize the path length to the closest workstation, instead of average path lengths to all workstations.

\mysubsubsection{Cluster Greedy Initialization.} 
% While the min-dist greedy initialization considers the movement of the robots on the sortation floor, 
This technique assigns chutes of the same destination closer to each other. This reduces the movement time of the human workers on the lower shipping floor, which in turn reduces the CLOSED time of chutes.
% number of timesteps the chutes must remain \emph{CLOSED}. 
The sooner the \emph{CLOSED} chutes can be \emph{OPEN} again to accept packages, the more candidate targets the robots have, thereby balancing the traffic on the sortation floor.

We design the above initialization methods such that the generated \chutemappings are guaranteed to be valid. With a population size of $\lambda$, our initial population consists of $\lambda - 2$ \chutemappings randomly sampled from the distribution of destinations as well as 2 \chutemappings from the min-dist and cluster greedy initializations, respectively. The pseudocode and detailed discussion of the initialization methods are included in Appendix~\ref{appen:ea-init}.
\subsection{Mutation} \label{subsec:ea-mutate}

Inspired by prior works~\cite{fontaine:gecco19,fontaine2021importance,zhangLayout23}, we mutate a selected \chutemapping by randomly replacing the destination mapped to each chute. Specifically, we randomly select $k$ chutes $\{c^{(1)},...,c^{(k)}\} \subseteq C$, where $k$ is sampled from a geometric distribution $P(X=k) = (1-p)^{k-1}p$ with $p=\frac{1}{2}$. For each selected chute $c^{(j)}, 1\leq j \leq k$, we replace the destination originally mapped to $c^{(j)}$ with a random one. If the mutated \chutemappings are invalid, we use the following MILP solver to repair them.

% To mutate a \chutemapping $\theta$, we first define $\psi: C \rightarrow D$ where $\forall c \in C$, $\psi(c) = d$ iff $c \in \theta(d)$. 
% Then , we randomly select $k$ chutes $\{c^{(1)},...,c^{(k)}\} \subseteq C$, where $k$ is sampled from a geometric distribution $P(X=k) = (1-p)^{k-1}p$ with $p=\frac{1}{2}$. For each selected chute $c^{(j)}, 1\leq j \leq k$, we change $\psi(c^{(j)}) \in $ to a random destination.

\subsection{MILP} \label{subsec:ea-milp}
Inspired by prior works~\cite{zhang:aiide2020,fontaine2021importance,zhangLayout23}, the MILP solver maps an arbitrary \chutemapping $\theta$ to a valid \chutemapping $\theta'$ with minimal modifications to $\theta$.  
Given $M$ chutes $\{c_1,...,c_M\}$ and $N$ destinations $\{d_1,...,d_N\}$ with their volumes $V = \{v_1,...,v_N\}$, we define $M \times N$ integer variables $e_{ij} \in \{0, 1\}, \forall i\in\{1,...,M\}, j\in\{1,...,N\}$, s.t. each variable $e_{ij}$ corresponds to whether we include the mapping from chute $i$ to destination $j$.
% The MILP attempts to make the least number of modifications to $\theta$ so that it becomes valid. 
Suppose $\theta$ is represented as $\{e_{ij}^{(0)} | i\in\{1,...,M\}, j\in\{1,...,N\} \}$, the objective minimizes the Hamming distance between $\theta$ and $\theta'$:

\begin{equation}
    \min \sum_{i=1}^M \sum_{j=1}^N |e_{ij} - e_{ij}^{(0)}|.
\end{equation}

To ensure $\theta'$ is valid, we include the following constraints:

\begin{align}
    \forall i \in \{1,\cdots,M\}, \sum_{j=1}^N e_{ij} = 1. \label{eq:chute-used-ones}\\
    \forall j \in \{1,\cdots,N\}, 1 \leq \sum_{i=1}^M e_{ij} \leq U_j \label{eq:bound-dest-es} \\
    \forall j \in \{1, \cdots, N\}, U_j = \max(1, \delta \cdot M \cdot \frac{v_j}{\sum_{j'=1}^{N} v_{j'}}). \label{eq:Uj}
\end{align}

\Cref{eq:chute-used-ones} ensures that each chute is assigned exactly one destination. \Cref{eq:bound-dest-es} ensures that each destination is assigned to $[1, U_j]$ chutes, where the upper bound $U_j$ is computed from \Cref{eq:Uj}. $U_j$ ensures an upper-bounded number of chutes for $d_j$ that is proportional to $v_j$.
% To maintain an approximately correct number of chutes for each destination, 
% We upper-bound the number of chutes assigned to each destination by a number that is proportional to the volume of the destination. 
$\delta$ is a hyperparameter that adjusts the upper bound. In our experiments, we set $\delta = 1.5$.
% so that the MILP solver has a wide range of valid \chutemappings to search for.

\section{Experimental Evaluation} \label{sec:exp}

\begin{figure*}
    \centering
    \includegraphics[width=0.9\linewidth]{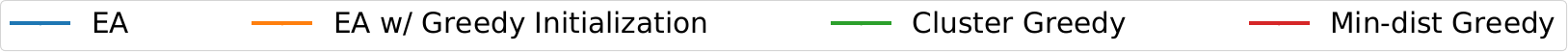}
    \begin{subfigure}{0.24\textwidth}
        \centering
        \includegraphics[width=1\textwidth]{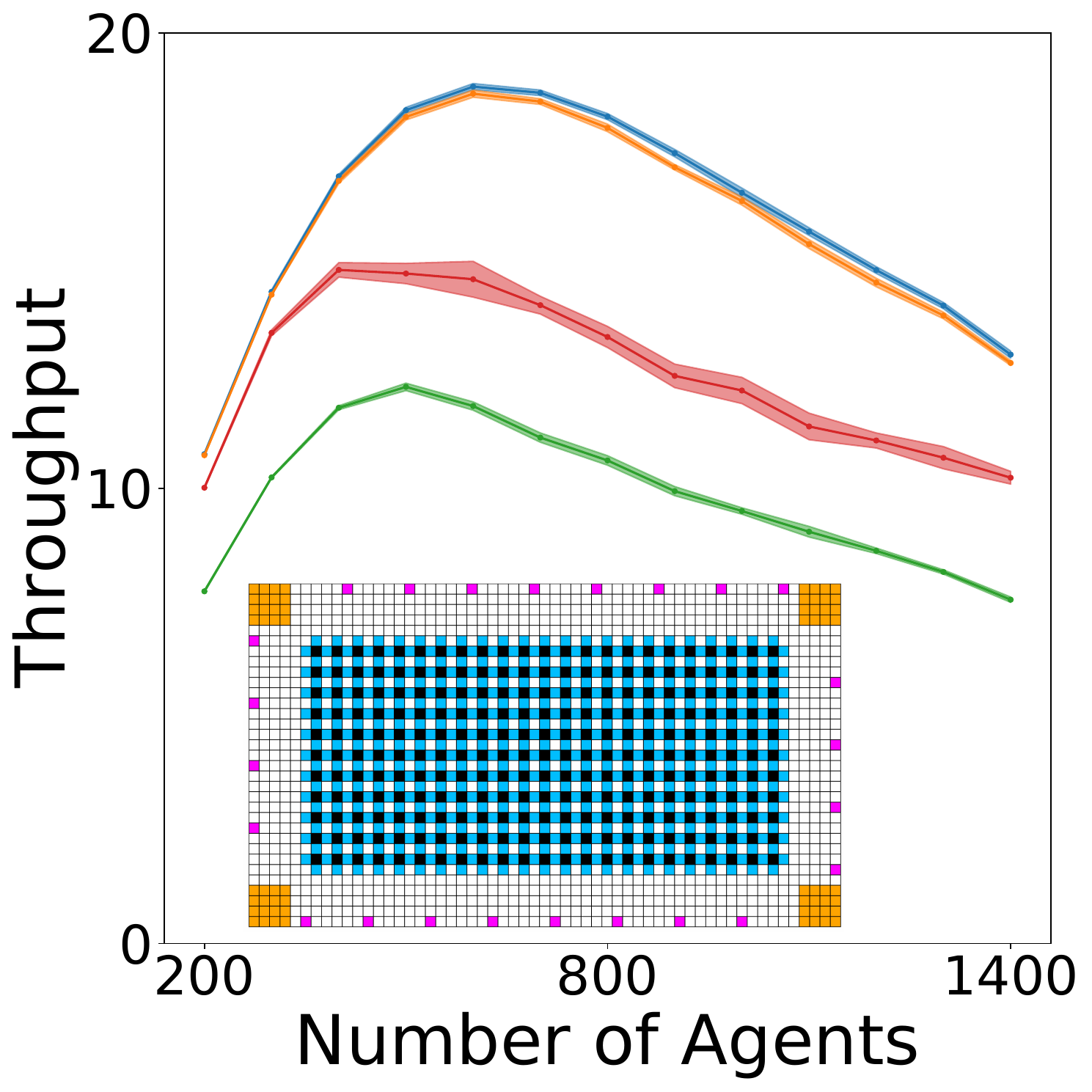}
        \caption{Setup 1}
        \label{fig:thr-agt-sortation-33-57}
    \end{subfigure}%
    \hfill
    \begin{subfigure}{0.24\textwidth}
        \centering
        \includegraphics[width=1\textwidth]{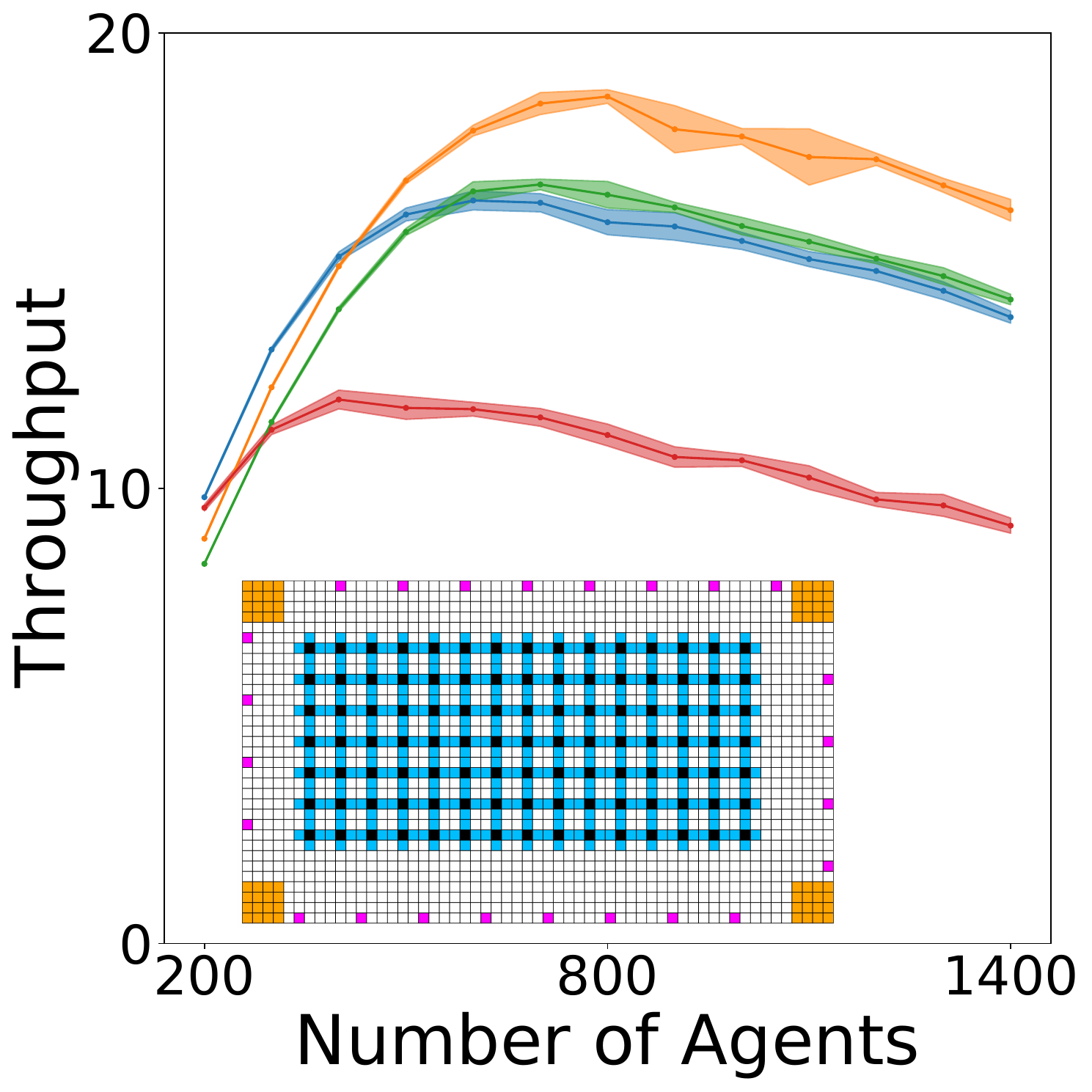}
        \caption{Setup 2}
        \label{fig:thr-agt-sortation-33-57-105}
    \end{subfigure}%
    \hfill
    \begin{subfigure}{0.24\textwidth}
        \centering
        \includegraphics[width=1\textwidth]{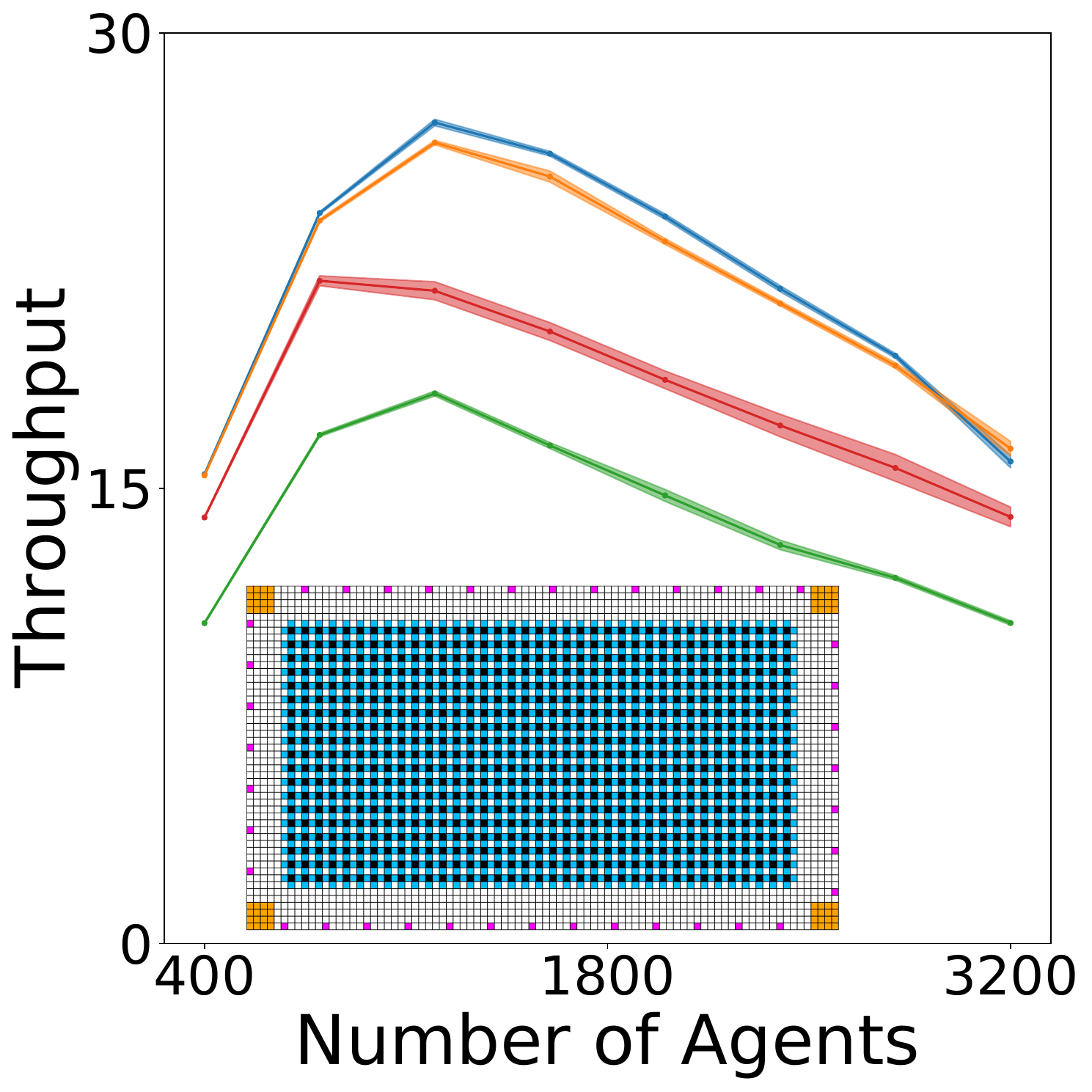}
        \caption{Setup 3}
        \label{fig:thr-agt-sortation-50-86}
    \end{subfigure}%
    \hfill
    \begin{subfigure}{0.24\textwidth}
        \centering
        \includegraphics[width=1\textwidth]{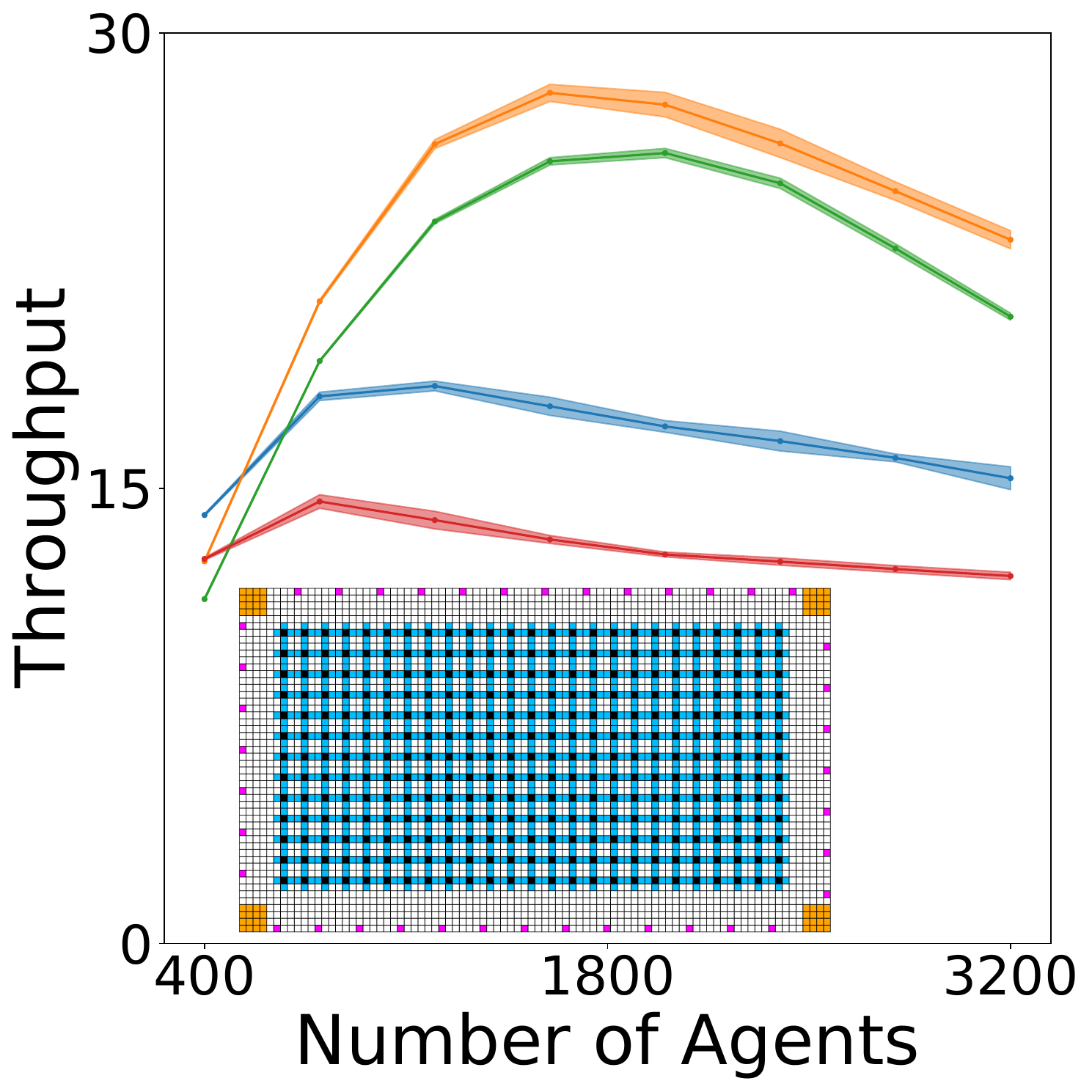}
        \caption{Setup 4}
        \label{fig:thr-agt-sortation-50-86-325}
    \end{subfigure}%
    \caption{Throughput with different numbers of robots. The \chutemappings are optimized with $N_a$ robots shown in \Cref{tab:exp-setup}. The solid line shows the average and the shaded area shows the 95\% confidence interval.}
    \label{fig:main-result}
\end{figure*}

In this section, we compare the optimized \chutemappings by our EA with various baselines.

\subsection{Experiment Setup}

\mysubsubsection{General Setups.}
\Cref{tab:exp-setup} shows the experimental setup. Columns 2 and 3 show information of the maps with their visualizations shown in \Cref{fig:main-result}. Notably, setups 1 and 2 and setups 3 and 4 have the same map sizes of 33 $\times$ 57 and 50 $\times$ 86, respectively, with different numbers of chutes. Column 4 shows the number of destinations for each map. We keep the ratio of $\frac{M}{N}$ similar across all maps to maintain a consistent capability of package sorting in all maps. Column 5 shows the number of robots used to optimize the \chutemappings.

\mysubsubsection{EA.} For all setups, we run EA with $N_{eval} = 10,000$, $\lambda = 100$, and $N_e = 5$, resulting in a total of $50,000$ simulations. We run each simulation for $N_T = 5,000$ timesteps. We keep the \chutemapping of the highest throughput as the final solution. To demonstrate the effect of initialization, we consider EA without the greedy initialization techniques (``EA'') and with the techniques (``EA w/ Greedy Initialization'').

\mysubsubsection{Baselines.} Since no prior works exist for the problem of TMO, we compare optimized \chutemappings by EA with the two greedy initialization methods, namely (1) Cluster Greedy and (2) Min-dist Greedy. The implementation and compute resources are included in Appendix~\ref{appen:compute}.
% In addition, to demonstrate the importance of initialization, we compare with a variant of EA that does not consider the two greedy initialization methods.

\begin{table}[!t]
    \centering
    % \resizebox{\linewidth}{!}{
    \begin{tabular}{ccccc}
    \toprule
    Setup & Map Size & $M$ & $N$ & $N_a$ \\
    \midrule
    1  & 33 $\times$ 57  & 253 & 99 & \multirow{2}{*}{600} \\
    2  & 33 $\times$ 57 & 105 & 41 & \\
    \cmidrule(lr){5-5}
    3  & 50 $\times$ 86  & 703 & 299 & \multirow{2}{*}{1200} \\
    4  & 50 $\times$ 86 & 325 & 138 & \\
    \bottomrule
    \end{tabular}
    % }
    \caption{
    Summary of the experiment setup. $M$ is the number of chutes, $N$ is the number of destinations, excluding the additional destination for recirculation, and $N_a$ is the number of robots used to optimize the \chutemappings.
    }
    \label{tab:exp-setup}
\end{table}

\subsection{Results}

\begin{table}[!t]
    \centering
    \resizebox{1\linewidth}{!}{
        \begin{tabular}{ccrr}
\toprule
Setup & TMO & Throughput ($\uparrow$) & Recirculation Rate ($\downarrow$) \\
\midrule
\multirow{4}{*}{1} & Cluster Greedy & $11.81 \pm 0.04$ & $2.19\% \pm 0.03\%$\\
  & Min-dist Greedy & $14.59 \pm 0.17$ & $2.90\% \pm 0.10\%$\\
  & \textbf{EA} & $\textbf{18.82} \pm \textbf{0.03}$ & $1.50\% \pm 0.02\%$\\
  & EA w/ Greedy Initialization & $18.67 \pm 0.03$ & $\textbf{1.34\%} \pm \textbf{0.03\%}$\\
\midrule
\multirow{4}{*}{2} & Cluster Greedy & $16.52 \pm 0.09$ & $4.87\% \pm 0.07\%$\\
  & Min-dist Greedy & $11.74 \pm 0.07$ & $10.20\% \pm 0.09\%$\\
  & EA & $16.32 \pm 0.09$ & $4.53\% \pm 0.18\%$\\
  & \textbf{EA w/ Greedy Initialization} & $\textbf{17.86} \pm \textbf{0.05}$ & $\textbf{3.16\%} \pm \textbf{0.11\%}$\\
\midrule
\multirow{4}{*}{3} & Cluster Greedy & $18.12 \pm 0.04$ & $1.36\% \pm 0.03\%$\\
  & Min-dist Greedy & $21.51 \pm 0.13$ & $2.26\% \pm 0.05\%$\\
  & \textbf{EA} & $\textbf{27.05} \pm \textbf{0.05}$ & $\textbf{1.20\%} \pm \textbf{0.03\%}$\\
  & EA w/ Greedy Initialization & $26.39 \pm 0.03$ & $1.33\% \pm 0.02\%$\\
\midrule
\multirow{4}{*}{4} & Cluster Greedy & $23.79 \pm 0.03$ & $3.21\% \pm 0.03\%$\\
  & Min-dist Greedy & $13.95 \pm 0.13$ & $8.89\% \pm 0.08\%$\\
  & EA & $18.37 \pm 0.07$ & $6.02\% \pm 0.06\%$\\
  & \textbf{EA w/ Greedy Initialization} & $\textbf{26.34} \pm \textbf{0.06}$ & $\textbf{2.98\%} \pm \textbf{0.05\%}$\\
\bottomrule
        \end{tabular}
    }
    \caption{Throughput and recirculation rate of the simulations with different \chutemappings. We report all results in the format of $x \pm y$ where $x$ is the average and $y$ is the standard error.}
    \label{tab:numerical-result}
\end{table}

\begin{figure*}
    \centering
    \begin{subfigure}{0.24\textwidth}
        \centering
        \includegraphics[width=1\textwidth]{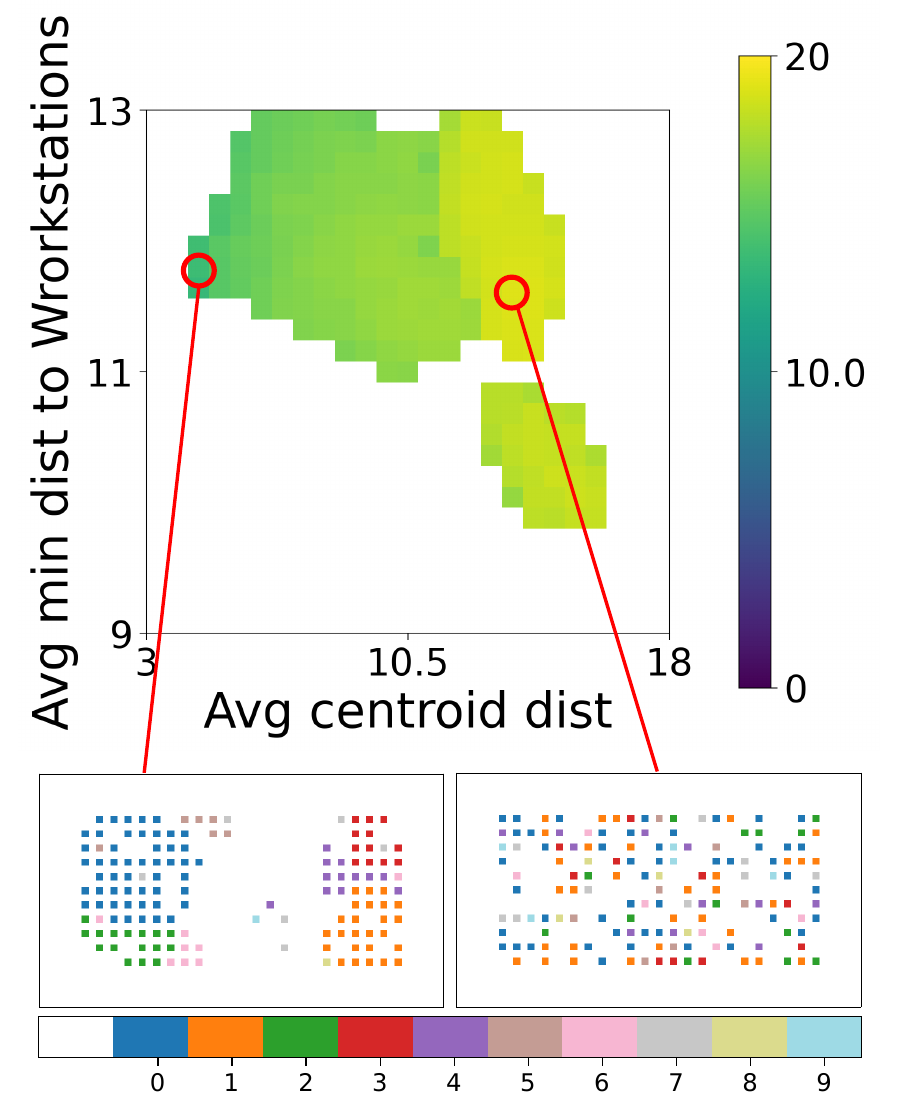}
        \caption{Setup 1}
        \label{fig:archive-sortation-33-57}
    \end{subfigure}%
    \hfill
    \begin{subfigure}{0.24\textwidth}
        \centering
        \includegraphics[width=1\textwidth]{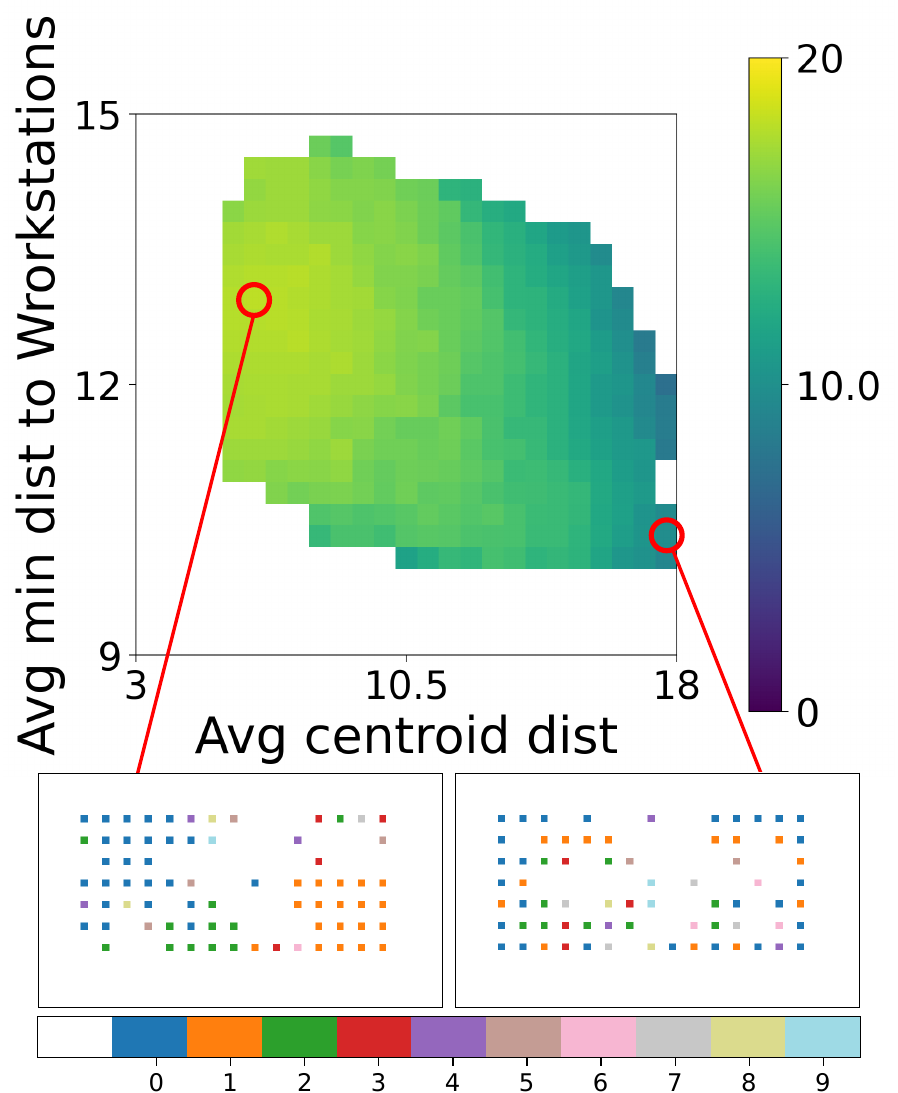}
        \caption{Setup 2}
        \label{fig:archive-sortation-33-57-105}
    \end{subfigure}%
    \hfill
    \begin{subfigure}{0.24\textwidth}
        \centering
        \includegraphics[width=1\textwidth]{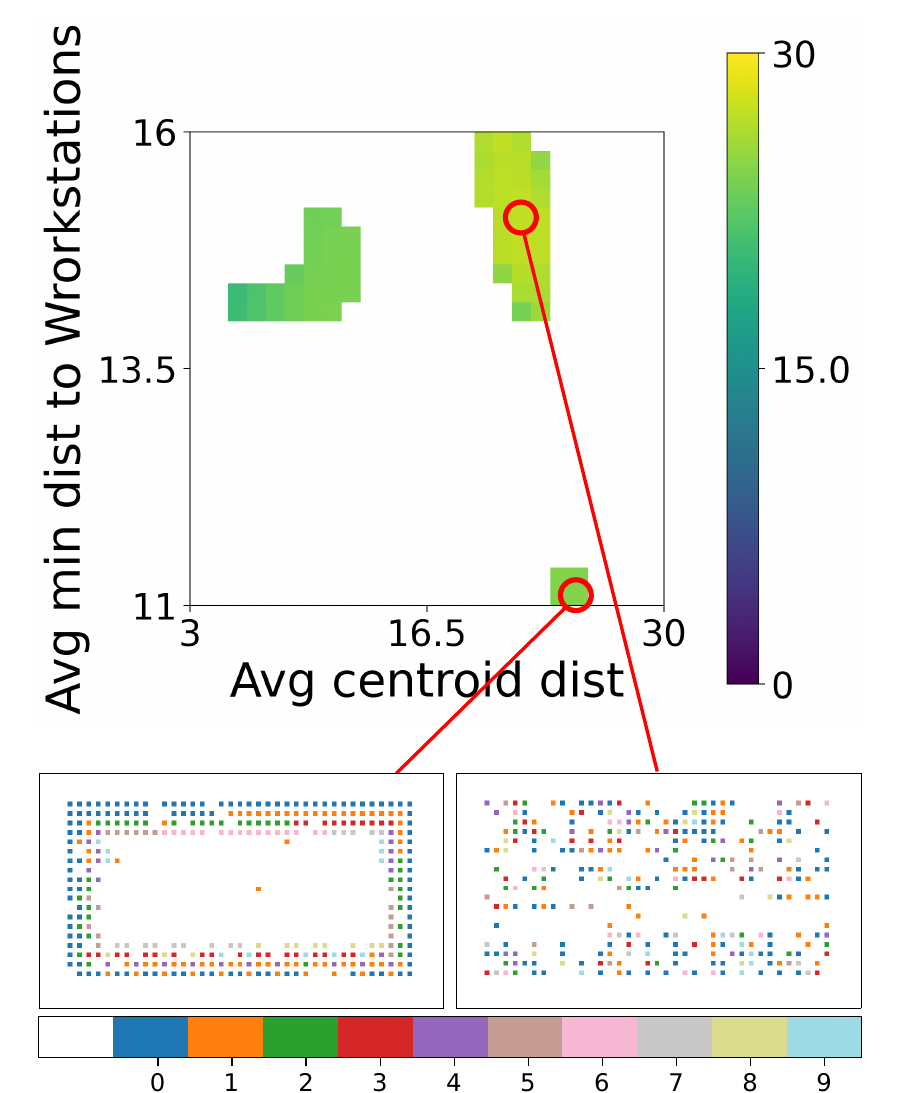}
        \caption{Setup 3}
        \label{fig:archive-sortation-50-86}
    \end{subfigure}%
    \hfill
    \begin{subfigure}{0.24\textwidth}
        \centering
        \includegraphics[width=1\textwidth]{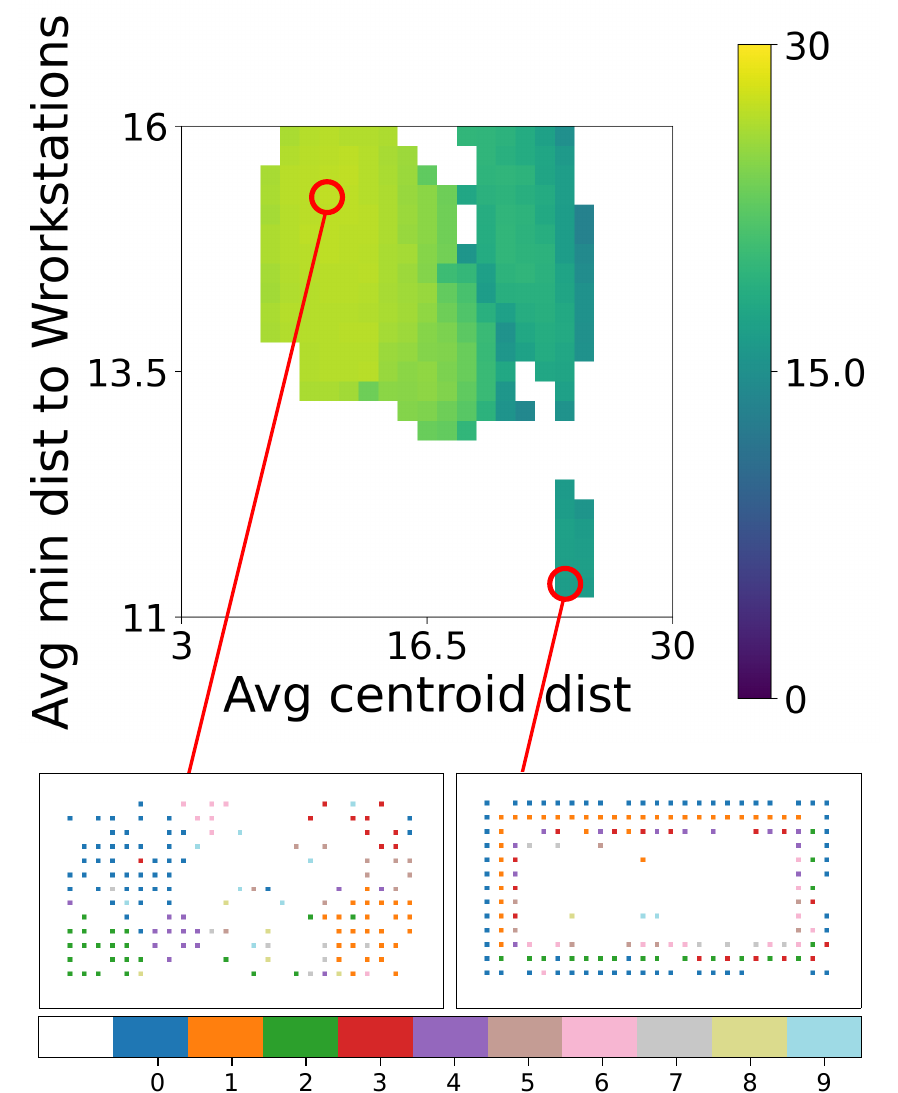}
        \caption{Setup 4}
        \label{fig:archive-sortation-50-86-325}
    \end{subfigure}%
    \caption{The archives of MAP-Elites in all setups. The two axes show the diversity measures while the color indicates the throughput. Two representative \chutemappings are shown for each setup. Each square in the \chutemappings represents a chute and the color represents the destination the chute is assigned to. The color bars underneath the \chutemappings shows the indices of the destinations. Smaller indices indicate larger volumes. For simplicity, we only show the top 10 destinations.}
    \label{fig:qd-result}
\end{figure*}

We first compare the variants of EA with the baselines. For each \chutemapping, we run 10 simulations with $N_a$ robots and report the numerical results in \Cref{tab:numerical-result}. We report throughput and \emph{recirculation rate}, defined as the ratio of recirculated packages. A lower recirculation rate indicates fewer delays in delivering packages.
In setups 1 and 3, the EA variants achieve higher throughput and lower recirculation rates. In setups 2 and 4, while EA w/ Greedy Initialization achieves the best throughput and recirculation rate, Cluster Greedy matches or outperforms EA. This is because setups 2 and 4 feature sortation floors with sparser chutes, giving robots more room to coordinate and alleviate traffic congestion. As a result, the bottleneck of throughput becomes the \emph{CLOSED} time of chutes. By placing chutes of the same destination together, the Cluster Greedy \chutemappings achieves reasonably high throughput in setups 2 and 4. By exploiting the Cluster Greedy \chutemappings as the initial solutions, EA w/ Greedy Initialization further improves the throughput in these setups.

The Min-dist Greedy \chutemappings outperform the Cluster Greedy ones in setups 1 and 3, where the traffic congestion is too severe because of dense chutes. Therefore, placing high-volume chutes close to workstations can take advantage of the empty space between chutes and workstations to coordinate the robots. However, surprisingly, using the Min-dist Greedy \chutemappings as the initial solutions does not allow EA w/ Greedy Initialization to yield better \chutemappings than EA in setups 1 and 3. 
Because Min-dist Greedy \chutemappings keep chutes \emph{CLOSED} for too long, they incur the highest recirculation rates in all setups and are likely less optimal than destination-sampled mappings.
We further demonstrate the performance of optimized \chutemappings by running simulations with various numbers of robots, each for 10 times, and show the results in \Cref{fig:main-result}. Our task mappings, though optimized with a particular number of robots, can be used with different numbers of robots, and the trend of the results aligns with our observation in \Cref{tab:numerical-result}.
% The trend of the results confirms our observation in \Cref{tab:numerical-result}.

\section{Empirical Analysis via QD Optimization}

The comparison between the optimized \chutemappings from EA and the greedy initialization techniques sparks an interesting question: what is the relationship between throughput and the objectives implicitly optimized by the greedy initialization techniques? In this section, we attempt to answer this question by performing empirical analysis using Quality Diversity (QD) algorithms.

\subsection{Quality Diversity Algorithms}

QD algorithms are a class of stochastic black-box optimization algorithms capable of simultaneously optimizing an objective and diversifying a set of diversity measure functions. 
% QD algorithms have wide applications in scenario generation~\cite{fontaine2021quality}, game content generation~\cite{fontaine2020illuminating,Zhang2021DeepSA}, reinforcement learning~\cite{batra2024proximal}, and layout optimization~\cite{zhangLayout23,ZhangNCA2023}.
We conduct the analysis by using MAP-Elites~\cite{mouret2015illuminating} to optimize the \chutemappings. MAP-Elites is identical to EA w/ Greedy Initialization except that it organizes the optimized \chutemappings in a discretized measure space defined by the diversity measure functions, referred to as an \emph{archive}. The goal of MAP-Elites is to search for the best \chutemapping in each cell of the archive. 
% referred to as \emph{elites}. 
% MAP-Elites has the same objective function as EA. 
% It discretizes the measure space defined by the diversity measure functions into evenly spaced cells, referred to as an \emph{archive}. Each cell in the archive corresponds to a range of values of the measure functions, and MAP-Elites tries to find the best solution in each cell, referred to as \emph{elites}. MAP-Elites maintains an archive and iteratively selects random elites from the archive, mutates them, and evaluates them to populate the archive. We use the same initialization and mutation techniques presented in \Cref{sec:tmo} in MAP-Elites. We use the same RSS simulator to evaluate the throughput and use it as the objective of MAP-Elites. 
% For the diversity measures, we formalize and use the objective optimized by Min-dist Greedy and Cluster Greedy initialization techniques as \emph{avg min dist to workstation (AMDW)} and \emph{avg centroid dist (ACD)}, respectively. To compute them, we first find the top 5\% of the destinations ranked by volumes as the high-volume destinations. For AMDW, we compute the average path length between chutes assigned to the high-volume destinations and their closest workstation. For ACD, we compute the average centroid distance of the chutes assigned to the high-volume destinations. 
For the diversity measures, we formalize and use the objectives optimized by Min-dist Greedy and Cluster Greedy initialization as \emph{avg min dist to workstation (AMDW)} and \emph{avg centroid dist (ACD)}, respectively. To compute them, we first identify the top 5\% of destinations ranked by volume as high-volume destinations. For AMDW, we calculate the average path length between chutes assigned to high-volume destinations and their nearest workstation. For ACD, we calculate the average centroid distance of the chutes assigned to these destinations.
% We implement MAP-Elites in Pyribs~\cite{pyribs}. 
% and use the same machines in \Cref{sec:exp} to conduct experiments.

\subsection{Experiment Setup and Results}

\mysubsubsection{Setup.}
We use the same setups in \Cref{tab:exp-setup}. For setups 1 and 2, we run MAP-Elites with $N_{eval} = 100,000$, $\lambda = 100$, $N_e = 5$, and $N_T = 5,000$. For setups 3 and 4, we use the same $\lambda$ and $N_e$ with a larger $N_{eval} = 200,000$, because they have larger search spaces. We set the resolution of the archives to be $25 \times 25$ in all setups.

\mysubsubsection{Results.} 
\Cref{fig:qd-result} shows the archives. The archives of setups 2 and 4 show similar trends with smaller ACD having higher throughput, aligning with our experimental results in \Cref{sec:exp}. The \chutemappings on the left of \Cref{fig:archive-sortation-33-57-105,fig:archive-sortation-50-86-325} show the optimal \chutemapping in the archive, with an average throughput of 17.98 and 27.16 by running 5 simulations, respectively. Both \chutemappings cluster chutes of the same destination together.
% , meaning that they are likely being mutated from the Cluster Greedy \chutemapping.
Since $N_{eval}$ of QD is much larger than EA, we obtain better throughput than the EA variants in \Cref{tab:numerical-result}. 
% With the same number of $N_{eval}$, the EA variants are superior.
On the other hand, in setups 1 and 3, larger ACD indicates better throughput. We conjecture that clustering chutes assigned to high-volume destinations together makes traffic too congested. The right \chutemappings of \Cref{fig:archive-sortation-33-57,fig:archive-sortation-50-86} show the optimal ones in the archive with an average throughput of 18.98 and 27.16. No clear patterns are shown in these \chutemappings, meaning that they are likely mutated from the initial \chutemappings being sampled from the distribution of destinations.
Notably, no clear relationship is shown between throughput and AMDW in all setups. We conjecture that AMDW does not fully capture the traffic pattern of the robots on the sortation floors, which depends on other factors such as floor layouts, coordination algorithms, and target assignment policies.

\section{Conclusion}

We introduce the \ChuteMapping Optimization (TMO) problem and present the first TMO approach for RSS, presenting a RSS simulator and showing that a high-quality \chutemapping can improve the throughput of RSS. 
% We demonstrate the complexity of a real-world RSS and present a simulator to evaluate the throughput. 
We use QD algorithms to conduct empirical analysis on the effect of different \chutemappings on throughput. 
Future works may attempt to make the RSS simulator more realistic or improve the sample efficiency of EA by using surrogate models~\cite{Zhang2021DeepSA}.
% include (1) making the RSS simulator more realistic by incorporating robot kinodynamics, (2) improving sample efficiency of EA by using surrogate models.

% Our work is limited in many ways. First, our RSS simulator has gaps with the real world. It ignores the kinodynamic constraints of the robots and simplifies the movement of human workers on the shipping floor. Second, our TMO method based on EA is computationally expensive, taking around 6 to 12 hours on machine (1) specified in Appendix~\ref{appen:compute}. Future works can explore using surrogate models~\cite{Zhang2021DeepSA} to improve the sample efficiency of EA. 
% Third, our work focuses on optimizing \emph{static} \chutemappings. Future work can explore optimizing \emph{dynamic} \chutemappings that change over time~\cite{shen_multi-agent_2023}.

\section*{Acknowledgment}

This work was partially supported by the National Science Foundation under grant \#2441629.

\bibliographystyle{IEEEtran}
\bibliography{reference}

\clearpage

\appendices

\section{EA Initialization} \label{appen:ea-init}

\begin{algorithm}[!ht]
\caption{Min-dist Greedy Initialization}
\label{alg:min-dist}
\LinesNumbered
\SetKwInput{KwInput}{Input}
\SetKwInput{KwOutput}{Output}
\SetKwProg{Fn}{Function}{:}{}

\KwInput{\\
$C = \{c_1,...,c_M\}$: Chutes sorted in ascending order w.r.t. path lengths to their corresponding closest workstations.\\
$D = \{d_1,...,d_N\}$, $V = \{v_1,...,v_N\}$: Destinations and their volumes sorted in descending order w.r.t. volumes.\\
% $d(c)$: function that computes path length from chute $c\in C$ to the closest workstation
}
\KwOutput{\Chutemapping $\theta$}

$\forall d \in D, \theta(d) = \{\}$ \label{alg:min-dist:init-cm} \\% initialize chutes of each destination to empty set

$i \gets 1$ \label{alg:min-dist:init-i} \\ % index of chutes

% $m_{r} \gets M$ \label{alg:min-dist:init-mr} \\ % remaining number of chutes

\For{$k \gets 1$ \KwTo $N$}
{\label{alg:min-dist:for}
    % Compute max number of chutes that should be mapped to current destination
    $m_{k} = \lfloor \frac{v_k}{\sum_{v\in V}v} \cdot M \rfloor + 1$ \label{alg:min-dist:max-chute} \\
    % Keep assigning until max chute is reached or there are not enough chutes for the rest of the destinations
    % $n_{r} \gets N - k$ \label{alg:min-dist:remain-dest}\\
    
    \While{$|\theta(d_k)| < m_{k} \And M - i + 1 > N - k$ }
    {\label{alg:min-dist:while}
        $\theta(d_k) \gets \theta(d_k) \cup \{c_i\}$ \label{alg:min-dist:add-chute} \\
        $i \gets i + 1$ \label{alg:min-dist:i+1} \\
        % $m_{r} \gets m_{r} - 1$ \label{alg:min-dist:mr-1} \\
    }
}

\Return $\theta$

\end{algorithm}

\begin{algorithm}[!ht]
\caption{Cluster Greedy Initialization}
\label{alg:cluster}
\LinesNumbered
\SetKwInput{KwInput}{Input}
\SetKwInput{KwOutput}{Output}
\SetKwProg{Fn}{Function}{:}{}
\SetKwFunction{centroid}{centroid}
\SetKwFunction{furthestFreeChute}{furthest\_free\_chute}
\SetKwFunction{closestFreeChute}{closest\_free\_chute}

\KwInput{\\
Chutes $C = \{c_1,...,c_M\}$.\\
$D = \{d_1,...,d_N\}$, $V = \{v_1,...,v_N\}$: Destinations and their volumes sorted in descending order w.r.t. volumes.\\
% $d(c)$: function that computes path length from chute $c\in C$ to the closest workstation
\centroid($\cdot$): Function to compute the centroid of a set of chutes \\
\furthestFreeChute($\cdot$): Function to compute the furthest unassigned chute to the current centroids.\\
\closestFreeChute($\cdot$): Function to compute the closest unassigned chute to a given set of chutes.
}
\KwOutput{\Chutemapping $\theta$}

$C_c \gets \{\}$ \label{alg:cluster:init-Cc} \\ % existing centroids
$m_r \gets M$ \label{alg:cluster:init-mr} \\ % Remain num of chutes

\For{$k \gets 1$ \KwTo $N$} { \label{alg:cluster:for}
    % max number of chutes can be assigned
    $m_{k} = \lfloor \frac{v_k}{\sum_{v\in V}v} \cdot M \rfloor + 1$ \label{alg:cluster:max-chute} \\
    % the exact number of chutes to be assigned
    % We must remain at least `N - k` chutes for the rest of the destinations (i.e. 1 chute for each remaining destination)
    $m_{a} = \min(m_{k}, \max(m_r - (N - k), 0))$ \label{alg:cluster:exact-chute} \\

    % Assign the first chute of d_k
    % If this is the first chute of the entire problem, put it on the top left corner. Otherwise, put it in a place that is furthest from the existing centroids.
    \uIf{$C_c$ is $\emptyset$}{ \label{alg:cluster:if-empty}
        $\theta(d_k) \gets \theta(d_k) \cup \{c_1\}$ \label{alg:cluster:ass-c1} \\
    }
    \Else { \label{alg:else}
        $c \gets $ \furthestFreeChute{$C_c,\theta, C$} \label{alg:cluster:find-far} \\
        $\theta(d_k) \gets \theta(d_k) \cup \{c\}$ \label{alg:cluster:add-far} \\
    }

    % Assign rest of the chutes for d_k one by one by finding the position that is closest to the current chutes of d_k
    \For{$j \gets 1$ \KwTo $m_{a} - 1$} { \label{alg:cluster:for-rest}
        $c \gets $ \closestFreeChute{$d_k, \theta, C$} \label{alg:cluster:find-close} \\
        $\theta(d_k) \gets \theta(d_k) \cup \{c\}$ \label{alg:cluster:add-close} \\
    }
    % update the number of remaining chutes
    $m_r \gets m_r - m_a$ \label{alg:cluster:mr} \\

    % update centroid
    $C_c \gets C_c \cup \{$\centroid{$\theta(d_k)$}$\}$ \label{alg:cluster:centroid} \\    
}

\Return $\theta$

\end{algorithm}

\Cref{alg:min-dist} describes the procedure of Min-dist Greedy Initialization. Since we prioritize assigning chutes closer to workstations to higher-volume destinations, we sort the chutes in ascending order by their path length to the closest workstation and sort the destinations by their volumes in descending order.
We initialize the \chutemapping (\Cref{alg:min-dist:init-cm}) and the index of the currently assigned chute (\Cref{alg:min-dist:init-i}). Then for each destination $d_k$, we first compute $m_k$ as the maximum number of chutes that can be assigned to $d_k$ in proportion to its volume, followed by the addition of 1 to ensure at least 1 chute is assigned to $d_k$ (\Cref{alg:min-dist:max-chute}). We then keep assigning the next available chute to $d_k$ until we reach $m_k$ or only one chute is left for each of the remaining destinations (\Cref{alg:min-dist:while,alg:min-dist:add-chute,alg:min-dist:i+1}).
This algorithm guarantees to generate a valid \chutemapping.

\Cref{alg:cluster} describes the procedure of Cluster Greedy Initialization. Similar to \Cref{alg:min-dist}, we sort destinations in descending order by their volumes. We first initialize the set of centroids of the chutes assigned to each destination (\Cref{alg:cluster:init-Cc}) and the number of remaining chutes (\Cref{alg:cluster:init-mr}). Then for each destination $d_k$ (\Cref{alg:cluster:for}), we compute $m_a$ as the exact number of chutes that should be assigned to $d_k$ (\Cref{alg:cluster:max-chute,alg:cluster:exact-chute}). 
% Specifically, $m_k$ computes the max number of chutes the same way as \Cref{alg:min-dist} and 
Intuitively, we try to assign as many chutes as possible while making sure that at least 1 chute is reserved for each remaining destination. We then start assigning chutes. We either assign a fixed chute $c_1$ on the top-left corner if $d_k$ is the first destination (\Cref{alg:cluster:if-empty,alg:cluster:ass-c1}), or we find the furthest unassigned chute from the current centroids (\Cref{alg:else,alg:cluster:find-far,alg:cluster:add-far}). 
% We select $c_1$ as the chute on the top-left corner. 
The rationale is to keep the chutes assigned to high-volume destinations as far as possible from each other, balancing the traffic. We then assign the rest of the chutes of $d_k$ one by one by looking for the closest unassigned chute to the chutes assigned to $d_k$ (\Cref{alg:cluster:for-rest,alg:cluster:find-close,alg:cluster:add-close}). This clusters chutes of the destination together, reducing the chute \emph{CLOSED} time. Finally, we update the number of remaining chutes (\Cref{alg:cluster:mr}) and the set of centroids (\Cref{alg:cluster:centroid}).

\section{Implementation and Compute Resource} \label{appen:compute}

\subsection{Implementation} 
We implement the EA and MAP-Elites in Pyribs~\cite{pyribs}, the MILP with IBM's CPLEX library~\cite{ibm_cplex}, and the RSS simulator based on the winning solution~\cite{Jiang2024Competition} of the 2023 League of Robot Runner Competition~\cite{chan2024lorr}, a lifelong MAPF competition focusing on coordinating a large fleet of robots in RMFS and RSS.

\subsection{Compute Resource}
Our experiments are not sensitive to runtime, allowing us to conduct experiments on three different machines: (1) a local machine with a 64-core AMD Ryzen Threadripper 3990X CPU and 192 GB of RAM, (2) a local machine with a 64-core AMD Ryzen Threadripper 7980X CPU and 256 GB of RAM, and (3) a local machine with a 16-core AMD 5950X CPU and 64 GB of RAM.

\end{document}